\documentclass[11pt]{article}

\usepackage[a4paper,margin=1in]{geometry}
\usepackage[T1]{fontenc}
\usepackage[utf8]{inputenc}
\usepackage{lmodern}
\usepackage{microtype}
\usepackage{amsmath,amssymb}
\usepackage{bm}
\usepackage{graphicx}
\usepackage{booktabs}
\usepackage{multirow}
\usepackage{array}
\usepackage{xcolor}
\usepackage{caption}
\usepackage{subcaption}
\usepackage{placeins}
\usepackage{algorithm}
\usepackage{algpseudocode}
\usepackage[numbers,sort&compress]{natbib}
\usepackage{authblk}
\usepackage[colorlinks=true,linkcolor=blue!55!black,citecolor=blue!55!black,urlcolor=blue!55!black]{hyperref}

\newcommand{\vbq}{\textsc{VBQ}}
\newcommand{\bits}[1]{\mbox{#1\,bit}}

\title{\textbf{Variable Bit-width Quantization:\\
Learning Per-Group Precision for ``Bigger-but-Smaller'' Language Models}}

\author[ ]{Hamish Ogilvy}
\affil[ ]{Algolia Research\\\texttt{hamish.ogilvy [at] algolia.com}}
\date{}

\begin{document}
\maketitle

\begin{abstract}
Low-bit quantization shrinks language models but treats precision as a single
global hyper-parameter: every weight is forced to the same bit-width. We
introduce \textbf{Variable Bit-width Quantization} (\vbq{}), a training-time
method in which each contiguous group of 64 weights \emph{learns} its own
resolution from $\{1,2,4,8\}$ bits through a Gumbel-Softmax relaxation, trained
jointly with the network by an alternating optimization that gives the precision
logits a clean, task-aligned signal. Across decoder-only transformers, \vbq{}
discovers a consistent and strongly heterogeneous allocation, \emph{within}
individual projection types and not merely across layers, that is impossible to
express with per-layer methods: $69\%$ of groups collapse to \bits{1}, the
language-model head averages $1.09$ bits, while the first MLP block keeps
$\sim\!2.5$ bits. The discovered pattern is stable enough to be frozen into a
fixed \emph{recipe} and reused without any further search. Using this recipe we
demonstrate a ``bigger-but-smaller'' regime: a 131M-parameter model at $1.82$
mean bits reaches validation perplexity $4.2$ on TinyStories, \emph{beating} a
55M FP16 model (PPL $4.4$) while using $3.8\times$ less storage, and the same
recipe lets a $1.46$B-parameter model on FineWeb-Edu match a $593$M FP16 control
at $\sim\!3.7\times$ less storage ($2.5\times$ more parameters, $\sim\!3.7\times$
smaller footprint). Measured as quality-per-byte, \vbq{}
is $3.9$--$8.4\times$ more efficient than FP16. Crucially, the recipe maps
directly to packed low-bit storage, so it \emph{also accelerates inference}:
with custom fused dequantize-and-multiply kernels, autoregressive decode, which
is memory-bandwidth bound, is faster at equal output, and because the benefit
is bandwidth-driven the speedup \emph{grows with scale} (parity at $131$M,
$1.9\times$ at $1.0$B, $4.7\times$ at $9$B on Apple silicon). We complement
perplexity with a distributional analysis (KL divergence and argmax-flip rate)
that reveals a striking mechanism: deeper layers progressively \emph{self-heal}
the quantization error injected by early layers. We delimit the method's scope
honestly: its win is a from-scratch, train-time phenomenon, and scaling the
search economically beyond $1.5$B parameters and adapting it to pretrained
checkpoints remain open. \vbq{} reframes precision as a learnable, non-uniform
resource and shows that spending a fixed bit budget unevenly is materially
better than spending it uniformly.
\end{abstract}

\section{Introduction}
The dominant lever for deploying large language models (LLMs) under tight memory
budgets is quantization: storing weights at fewer than 16 bits. Almost all
deployed schemes (post-training methods such as GPTQ~\citep{frantar2023gptq} and
AWQ~\citep{lin2024awq}, and quantization-aware training such as
BitNet~\citep{wang2023bitnet,ma2024bitnet158}) fix a \emph{single} bit-width for
(almost) all weights. The schemes that do vary it still fix the allocation by a
predetermined rule rather than learning it. Production toolchains such as
Unsloth's dynamic GGUF quants~\citep{unsloth2025dynamic} keep salient layers at
higher precision as a per-layer, post-training heuristic; NVIDIA's
NVFP4~\citep{nvidia2025nvfp4} goes finer, mixing precision \emph{within} a tensor
through 16-element micro-blocks (and reserving a few hand-picked layers for higher
precision), and is used both for post-training quantization and for 4-bit
pretraining~\citep{nvidia2025nvfp4pretrain}. In every case the bit allocation
follows a fixed format or heuristic; it is not a per-group choice the
optimization \emph{learns} during training.

This is a convenient, yet suboptimal modeling choice. Weights in an LLM are not equally important:
attention output projections tolerate aggressive compression, while the first
feed-forward block and a small set of frequent output-vocabulary rows are far
more sensitive. A method that could \emph{allocate} bits where they matter, at
fine granularity and decided by the optimization itself, should dominate any
uniform assignment on the quality/size frontier.

A second observation motivates \emph{how much} precision is actually needed, and
is what set this work in motion. In production neural retrieval, the neural
hashing that compresses embedding vectors in Algolia's search
stack~\citep{searchio2022neuralhashing} was originally trained on FP32 vectors and
compressed them roughly $10\times$, to about $3.2$ bits per dimension, with a
change in ranking quality (nDCG) smaller than the noise of the embedding model
itself; we later relaxed this to $4$ bits, but pushing \emph{below} that range
degraded quality quickly. We had treated this as a practical rule of
thumb: a trained representation seems to carry on the order of $3$--$4$ useful
bits per value, a limit we hit in production, and one that has since surfaced
repeatedly in the LLM literature. Apple's on-device foundation model is
independent, production-scale evidence of the same limit: it compresses to a
\emph{mixed} 2-/4-bit scheme averaging $3.7$ bits-per-weight to \emph{match} the
uncompressed model (and $3.5$ bits without significant quality
loss)~\citep{apple2024fm}. The same threshold recurs in
\citet{dettmers2023case}, who find $4$-bit Pareto-optimal in their $k$-bit
inference scaling laws. That the largest deployers either still pay for 16-bit
weights or accept lossy quantization, yet a carefully mixed scheme reaches
$\sim\!4$ bits without measurable quality loss, is what made the opportunity
concrete: there was clearly room to operate near $4$ bits losslessly; the open
question was \emph{how} to get there. We read this convergence across retrieval,
on-device LLMs, and scaling-law analyses as an empirical floor: below roughly
$3.5$--$4$ bits per value, information is genuinely lost. \vbq{}'s
``bigger-but-smaller'' result is consistent with this floor, but with a crucial
twist about \emph{how} to reach it. The way to hit the floor is \emph{not} to
round every weight to four bits: uniform 4-bit quantization forces a lossy
approximation of weights that were never trained for it. It is to spend a
comparable total budget \emph{differently}: distribute it across more,
lower-precision weights (in our recipe, an average of $\sim\!1.8$ bits) so the network learns a
representation that lives natively at low bits. The information budget is
conserved; the layout is what changes, and, as we show, layout is decisive.

We propose \textbf{Variable Bit-width Quantization} (\vbq{}). Each group of 64
contiguous weights is given a categorical variable over the candidate
bit-widths $\{1,2,4,8\}$, relaxed with Gumbel-Softmax~\citep{jang2017gumbel,
maddison2017concrete} so it is differentiable. The central difficulty is that the
cross-entropy gradient, flowing through the quantizer, overwhelms the
bit-budget penalty by two-to-three orders of magnitude, freezing the precision
logits at their initialization. We resolve this with two design choices:
\emph{gradient isolation} (the reconstructed weights are detached when computing
the bit-selection path, so cross-entropy cannot back-propagate into the precision
logits) and an \emph{alternating optimization} (weight steps and precision steps
are interleaved, in the spirit of DARTS~\citep{liu2019darts}). Together these
let the logits actually move and commit.

Our key empirical findings are:
\begin{enumerate}
\itemsep2pt
\item \textbf{A learnable, heterogeneous precision hierarchy emerges}
(Section~\ref{sec:discovery}). The network drives $69\%$ of groups to \bits{1}
and reserves $4$/\bits{8} for a minority, producing a $1.78$-bit mean. The
allocation is heterogeneous \emph{within} projection types, structurally
impossible for per-layer mixed-precision methods.
\item \textbf{The recipe is the product, not the search}
(Section~\ref{sec:recipe}). Bit assignments lock in by $\sim\!20\%$ of training;
freezing them into a fixed recipe (uniform \bits{2} blocks, a $4$-bit first
block, and a frequency-tiered language-model head averaging $1.18$ bits)
reproduces the adaptive result with no Gumbel machinery at deployment.
\item \textbf{Bigger-but-smaller} (Section~\ref{sec:bigger}). A 131M model at
$1.82$ bits (29\,MB) \emph{beats} a 55M FP16 model (109\,MB) on perplexity, and
the same recipe lets a $1.46$B model ($2.5\times$ more parameters) match a
$593$M FP16 control on FineWeb-Edu at $\sim\!3.7\times$ less storage.
\item \textbf{Depth self-heals quantization error}
(Section~\ref{sec:kld}). A KL-divergence decomposition shows that the
representational gap injected by the early 4-bit block is progressively reduced
by later 2-bit blocks; the per-position argmax-flip rate vs.\ FP16 falls
monotonically from $71\%$ to $28\%$ across depth.
\item \textbf{The compression is real at inference time}
(Section~\ref{sec:inference}). The recipe maps to packed $\{1,2,4,8\}$-bit
storage that our fused dequantize-and-multiply kernels consume directly, never
materializing FP16 weights. Because autoregressive decode is memory-bandwidth
bound, the storage win becomes a speed win that grows with model size: from
parity at $131$M to $4.7\times$ faster decode at $9$B, at bit-exact agreement,
on Apple silicon.
\end{enumerate}

\vbq{}'s win is a \emph{from-scratch, train-time} phenomenon rather than a free
lunch for compressing arbitrary pretrained checkpoints; we make this scope
explicit and report the relevant open problems (chiefly economical scaling of
the search beyond $1.5$B parameters) in Section~\ref{sec:limits}.

\section{Related Work}
\paragraph{Low-bit and ternary LLMs.}
BitNet~\citep{wang2023bitnet} and BitNet~b1.58~\citep{ma2024bitnet158} train
transformers with 1-bit / ternary weights from scratch, fixing the same
precision everywhere. ParetoQ~\citep{liu2025paretoq} unifies 1- to 4-bit QAT and
finds that most of the training budget is best spent in full precision with a
short low-bit fine-tune, a different regime from our purely from-scratch QAT.
Low-bit models have also gained traction at the deployment frontier: the
Bonsai family~\citep{prismml2026bonsai} ships open-weight 1-bit and ternary
(1.58-bit) models up to $8$B parameters with strong reported accuracy-per-byte
for on-device inference, obtained by quantization-aware training to a single
\emph{uniform} low-bit format. We note it as evidence of real appetite for the
``bigger-but-smaller'' regime; \vbq{} differs in \emph{learning} a heterogeneous
per-group allocation, of which a uniform budget is a special case, so a
successfully-learned non-uniform allocation can only match or improve on a
uniform one at equal bytes (we do not attempt a head-to-head comparison, as the
scale, tokenizer, and evaluation suite differ).
\vbq{} differs from all of these in that the bit-width is \emph{not} a global
constant but a per-group learned quantity.

\paragraph{Post-training quantization (PTQ).}
GPTQ~\citep{frantar2023gptq}, AWQ~\citep{lin2024awq},
SmoothQuant~\citep{xiao2023smoothquant}, QuIP\#~\citep{tseng2024quip} and
AQLM~\citep{egiazarian2024aqlm} compress an already-trained model using
calibration data, second-order information, or learned codebooks.
LLM.int8()~\citep{dettmers2022llmint8} keeps a few outlier channels in higher
precision via a fixed threshold. Production toolchains increasingly ship
\emph{per-layer} dynamic mixed precision in this PTQ regime: Unsloth's dynamic
GGUF quants~\citep{unsloth2025dynamic} leave salient layers at higher precision
via an importance heuristic while pushing the rest to low bit-widths: a
post-training, per-tensor analogue of the per-group allocation \vbq{} learns
during training. These are complementary to \vbq{}: they operate
\emph{after} training and at (mostly) fixed, per-layer bit-widths, whereas we
learn the allocation \emph{during} training at per-group granularity.

\paragraph{Mixed-precision search.}
Learning or searching per-layer bit-widths is well studied for CNNs:
HAQ~\citep{wang2019haq} (RL), HAWQ~\citep{dong2019hawq} (Hessian),
and the differentiable DNAS / EdMIPS family~\citep{wu2018mixed,cai2020edmips}.
For LLMs, MixLLM~\citep{zheng2024mixllm} and SliM-LLM~\citep{huang2024slimllm}
allocate precision across output features or groups, but as
\emph{post-training} salience heuristics. The closest in spirit is the
differentiable, Gumbel-based CNN search of \citet{wu2018mixed}; \vbq{} brings
this idea to decoder-only LLMs, at \emph{per-group} (64-weight) granularity, with
gradient isolation and alternating optimization that are necessary to make
training-time selection work at this scale. To our knowledge, no prior work
learns per-group bit-widths jointly during training for LLMs.

\paragraph{Concurrent work on differentiable quantization.}
Two recent papers apply differentiable discrete relaxations to LLM precision in
\emph{post-training} settings, and clarify what is distinct about \vbq{}.
GSQ~\citep{dadgarnia2026gsq} uses a Gumbel-Softmax relaxation to jointly learn
per-coordinate scalar-grid assignments and per-group scales of an
\emph{already-trained} model, matching trellis-quantized accuracy while staying
in a deployment-friendly scalar format; it optimizes a layer-wise
reconstruction loss rather than the task loss, and selects grid points rather
than power-of-two bit-widths. dMX~\citep{franco2026dmx} learns \emph{per-layer}
floating-point (OCP MX) formats via a continuous offset annealed to discrete
formats. \vbq{} is complementary on three axes: it learns \emph{integer}
bit-widths at \emph{per-group} (64-weight) granularity, \emph{from scratch}
against the task loss, and distills the result into a fixed recipe; the shared
lesson across all three is that annealed differentiable relaxations are an
effective way to make discrete precision decisions trainable.

\paragraph{Scaling laws and the practical precision floor.}
\citet{kumar2025precision} and \citet{cao2024mixedscaling} formalize how
parameter count trades against precision, and \citet{dettmers2023case} study
$k$-bit inference scaling and find $4$-bit Pareto-optimal. This $\sim\!4$-bit
regime is borne out in production: Apple's on-device foundation
model~\citep{apple2024fm} compresses to a \emph{mixed} 2-/4-bit scheme averaging
$3.7$ bits-per-weight that matches the uncompressed model (and $3.5$ bits without
significant quality loss), a deployed mixed-precision system landing at exactly
the precision floor that also motivates \vbq{} (Section~\ref{sec:method}).
Hardware is converging on the same regime: NVIDIA's NVFP4~\citep{nvidia2025nvfp4}
is a $4$-bit floating-point format with two-level micro-block scaling (a
per-$16$-value FP8 scale and a per-tensor FP32 scale) that holds under $1\%$
accuracy loss versus FP8 on some models, but it fixes a single $4$-bit format
across the network rather than learning where precision is needed. Our
``bigger-but-smaller'' results are a concrete instantiation of this trade in the
per-group mixed-precision setting: rather than round each weight to the $\sim\!4$-bit
floor, spend a comparable budget on more, lower-precision parameters.

\paragraph{Evaluation beyond perplexity.}
\citet{dutta2024accuracy} show that accuracy alone hides large behavioral
changes from compression, and advocate KL divergence and \emph{flips} (answers
changing) as distance metrics. We adopt both to audit \vbq{}
(Section~\ref{sec:kld}).

\section{Method}\label{sec:method}

\subsection{Per-group quantization}
A weight matrix $W$ is partitioned into contiguous groups of $G=64$ weights.
Each group $g$ has a symmetric absmax scale $s_g$ and a bit-width $b_g$. For a
target bit-width $b$, the quantizer is
\begin{equation}
s_g = \frac{\max_{i\in g}|w_i|}{2^{\,b-1}-1},\qquad
\hat{w}_i = s_g \cdot \mathrm{clip}\!\Big(\mathrm{round}(w_i/s_g),\,
-(2^{\,b-1}-1),\,2^{\,b-1}-1\Big).
\end{equation}
The 1-bit case is handled as sign with a per-group absmean scale,
$\hat{w}_i = \mathrm{sign}(w_i)\cdot\mathrm{mean}_{j\in g}|w_j|$. Gradients pass
through the round/clip via the straight-through estimator~\citep{bengio2013ste}.

\paragraph{Why group size matters.}
With $G{=}1$ the scheme is degenerate: for any single weight,
$\hat{w}=\mathrm{round}(w/s)\cdot s = w$ regardless of $b$, so all bit-widths
produce identical outputs and there is nothing to choose. A shared per-group
scale ($G{=}64$) is what creates a genuine quality/precision trade-off and,
empirically, the heterogeneous allocation we observe (Section~\ref{sec:discovery}).

\subsection{Differentiable bit selection}
Each group carries logits $\bm{\ell}_g\in\mathbb{R}^{K}$ over the $K{=}4$
candidates $\mathcal{B}=\{1,2,4,8\}$. During the selection path we draw a
Gumbel-Softmax sample $\bm{\pi}_g=\mathrm{softmax}((\bm{\ell}_g+\bm{e}_g)/\tau)$
with Gumbel noise $\bm{e}_g$ and temperature $\tau$ annealed from $5.0$ to
$0.5$, and form the soft-quantized weight as a convex combination of the
candidate reconstructions,
$\hat{w}^{\text{soft}}_i = \sum_{k} \pi_{g,k}\,\hat{w}^{(\mathcal{B}_k)}_i$.
The expected bit cost of the network is
$\bar{b} = \frac{1}{N}\sum_g \sum_k \pi_{g,k}\,\mathcal{B}_k$, and the training
objective is $\mathcal{L} = \mathcal{L}_{\text{CE}} + \lambda\,\bar{b}$.

\paragraph{What the objective optimizes (and why \vbq{} is not lossless at fixed size).}
$\mathcal{L}$ is a \emph{penalty} (Lagrangian) objective, not a
quality-constrained one: $\lambda$ is an exchange rate that trades cross-entropy
for bits, and each group is compressed until the marginal quality cost of one
fewer bit exceeds $\lambda$. We deliberately choose $\lambda$ so the mean lands at
$\sim\!1.8$ bits, \emph{below} the $3.5$--$4$-bit information floor discussed in
the introduction. At a \emph{fixed} parameter count \vbq{} is therefore
\emph{expected} to be worse than an FP16 model: pushing precision below the floor
genuinely discards information, and this is by design, not a mis-tuned selection.
The capacity is recovered not by squeezing each weight but by \emph{layout}:
spending the same total bit budget across more, lower-precision weights that the
network learns to use natively (the ``bigger-but-smaller'' result). This depends
on weights \emph{co-adapting} to their assigned bit-widths \emph{during} training;
it cannot be obtained by lowering the bits-per-weight of an already-trained model,
which merely forces a lossy approximation of weights that were never trained for
it. In other words, \vbq{} does not compress a trained network to fewer bits; it
starts from a more efficient bit \emph{layout} chosen for the target accuracy and
storage.

\subsection{Gradient isolation and alternating optimization}\label{sec:altopt}
The naive objective does not work: the cross-entropy gradient reaching
$\bm{\ell}_g$ through $\hat{w}^{\text{soft}}$ is $\sim\!16\times$ larger than the
penalty gradient \emph{at every temperature}, and Adam's normalization
entrenches whichever logit started largest. The logits never move.

We make two changes. (i) \textbf{Gradient isolation}: in the bit-selection path
the candidate reconstructions are detached from the weight values
($\hat{w}^{(\mathcal{B}_k)} = \texttt{detach}(\cdot)$), so cross-entropy cannot
back-propagate into $\bm{\ell}_g$; only the bit penalty trains the precision
decision. (ii) \textbf{Alternating optimization} (cf.\ DARTS~\citep{liu2019darts}):
we interleave a \emph{weight step} (update $W$ on $\mathcal{L}_{\text{CE}}$ with
logits frozen) and a \emph{logit step} (update $\bm{\ell}$ on
$\mathcal{L}_{\text{CE}}+\lambda\bar b$ with $W$ frozen). This gives the logits a
clean, task-aligned signal about how precision affects loss, rather than the
contaminated gradient that froze them. With both changes the bit distribution
collapses from near-uniform to a committed (entropy $<\!1\%$) heterogeneous
allocation. Algorithm~\ref{alg:vbq} makes the two steps precise.

\begin{algorithm}[t]
\small
\caption{\vbq{} alternating optimization (one logit update every $m$ weight steps)}
\label{alg:vbq}
\begin{algorithmic}[1]
\Require weights $W$, precision logits $\bm{\ell}$, candidates
$\mathcal{B}=\{1,2,4,8\}$, penalty $\lambda$, temperature $\tau_t$, interval $m$
\For{each training step $t$}
  \State $\hat{w}^{(\mathcal{B}_k)} \gets \texttt{quant}(W, \mathcal{B}_k)$ for each $k$ \Comment{per-group reconstructions, Eq.~(1)}
  \If{$t \bmod m \neq 0$} \Comment{\textbf{weight step}: logits frozen}
    \State $\bm{\pi} \gets \texttt{detach}\big(\mathrm{GumbelSoftmax}(\bm{\ell}, \tau_t)\big)$
    \State $\hat{w} \gets \sum_k \pi_k\,\hat{w}^{(\mathcal{B}_k)}$;\quad
           $\;W \gets W - \eta\,\nabla_{W}\,\mathcal{L}_{\text{CE}}(\hat{w})$
  \Else \Comment{\textbf{logit step}: weights frozen, $\hat{w}^{(\mathcal{B}_k)}$ detached}
    \State $\bm{\pi} \gets \mathrm{GumbelSoftmax}(\bm{\ell}, \tau_t)$;\quad
           $\bar b \gets \tfrac{1}{N}\sum_g\sum_k \pi_{g,k}\,\mathcal{B}_k$
    \State $\bm{\ell} \gets \bm{\ell} - \eta\,\nabla_{\bm{\ell}}\big(\mathcal{L}_{\text{CE}}(\texttt{sg}[\hat{w}]) + \lambda\,\bar b\big)$
  \EndIf
\EndFor
\State \textbf{return} hardened bit-widths $b_g = \mathcal{B}_{\arg\max_k \ell_{g,k}}$
\end{algorithmic}
\end{algorithm}

Here $\texttt{sg}[\cdot]$ ($\equiv\texttt{detach}$) is the stop-gradient that
implements gradient isolation: in the logit step the reconstructions are held
constant so the only path into $\bm{\ell}$ is through $\bar b$ and the explicit
dependence of $\bm{\pi}$ on $\bm{\ell}$.

\subsection{From discovery to a fixed recipe}\label{sec:recipe}
Because the allocation locks early (Section~\ref{sec:discovery}), production
training does not need the Gumbel machinery at all. We distill the discovered
pattern into a fixed \textbf{\vbq{} recipe}:
\begin{itemize}
\itemsep1pt
\item First block MLP (\texttt{gate/up/down}): \bits{4} (most sensitive layer).
\item All other attention and MLP projections: \bits{2}.
\item Language-model head: frequency-tiered per-row mixed precision. We count
token frequencies in the training corpus, rank the vocabulary in descending
order of frequency, and assign bit-widths by rank percentile: top $0.5\%$ of
tokens at \bits{8}, next $1.5\%$ at \bits{4}, next $10\%$ at \bits{2}, bottom
$88\%$ at \bits{1} (mean $1.18$ bits). These four cutoffs are not tuned per run;
they are chosen once to reproduce the per-row allocation that the adaptive
search discovers for the head (Section~\ref{sec:discovery}).
\end{itemize}
The recipe is the single most transferable artifact of the method: it requires
only a token-frequency table for the head (a single pass over the corpus) and a
one-line projection-name map for a new architecture.

\section{Experimental Setup}
\paragraph{Models and data.}
The discovery and ablation studies use nanoGPT-style decoder-only transformers
(RoPE~\citep{su2024roformer}, SwiGLU~\citep{shazeer2020glu}, weight-tied head)
trained on TinyStories~\citep{eldan2023tinystories} (200M tokens, GPT-2
tokenizer~\citep{radford2019gpt2}). Scale-up experiments use
FineWeb-Edu~\citep{penedo2024fineweb} (2B tokens, Qwen
tokenizer~\citep{yang2024qwen3}). Sizes range from 55M to 1.46B parameters.

\paragraph{Training.}
AdamW ($\beta=(0.9,0.95)$, weight decay $0.1$), cosine schedule
$3\!\times\!10^{-4}\!\to\!3\!\times\!10^{-5}$ with warmup, gradient clip $1.0$.
TinyStories runs use $35$K iterations. Storage is reported as the packed weight
size at the realized mean bit-width (embedding/head included); ``FP16'' baselines
store every weight in 16 bits. We report validation perplexity (PPL), and for the
distributional study, mean token-level KL divergence to an FP16 reference and the
argmax-flip rate.

\section{Results}

\subsection{A learned, heterogeneous precision hierarchy}\label{sec:discovery}
\begin{figure}[t]
  \centering
  \includegraphics[width=\textwidth]{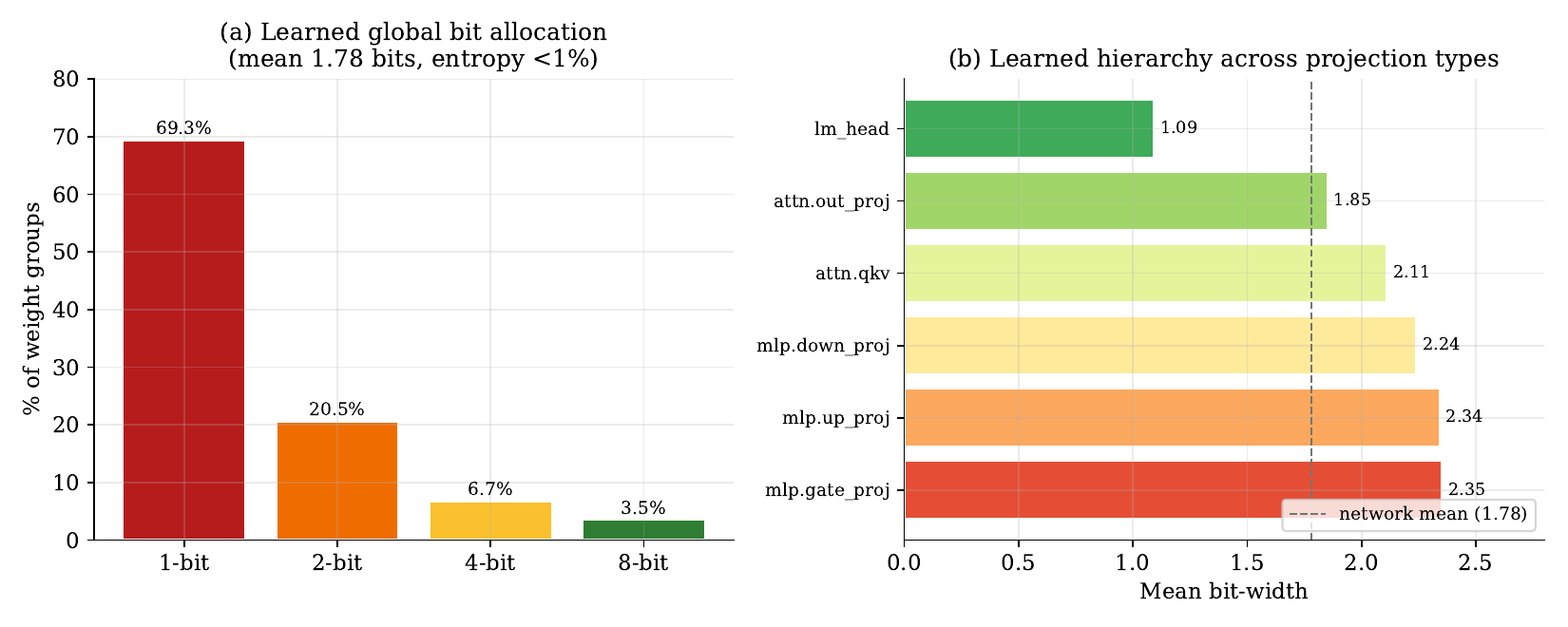}
  \caption{\textbf{\vbq{} learns a heterogeneous precision allocation.}
  (a) Global distribution of learned bit-widths (mean $1.78$, selection entropy
  $<1\%$). (b) Mean bit-width per projection type: the language-model head
  collapses to near-binary while the first-block MLP stays $>2$ bits. The
  hierarchy is discovered, not hand-set.}
  \label{fig:alloc}
\end{figure}

The adaptive pilot (58M, 8 layers, TinyStories) converges to a mean of
$1.78$ bits with a sharply heterogeneous distribution
(Figure~\ref{fig:alloc}a): $69.3\%$ of groups choose \bits{1}, $20.5\%$
\bits{2}, $6.7\%$ \bits{4}, $3.5\%$ \bits{8}, at near-zero selection entropy. The
allocation follows a stable \emph{projection-type hierarchy}
(Figure~\ref{fig:alloc}b): the language-model head averages $1.09$ bits
($92.5\%$ of its rows are pure \bits{1}), attention output projections are the
cheapest in-block projection, and the SwiGLU gate/up projections of the
\emph{first} block are the hungriest (up to $2.5$ bits with $\sim\!10\%$ \bits{8}
groups). Crucially, the heterogeneity exists \emph{within} a projection type, not
only across layers, a degree of freedom no per-layer method can use.

The head's allocation is semantically interpretable: $4.8\%$ of the
$1000$ most-frequent vocabulary rows contain \bits{8} groups versus $0.06\%$ of
the rarest, and the high-precision rows are dominated by function words and
punctuation. The same frequency structure is what motivates the recipe's
frequency-tiered head (Section~\ref{sec:recipe}). Finally, the allocation
\emph{locks} by $\sim\!10$K of $50$K steps; the remaining $80\%$ of training only
tunes weights within fixed precision, empirical justification for prescribing
the recipe directly.

\paragraph{The discovery transfers to a different architecture (DeltaNet).}
To test whether this hierarchy is an artifact of standard transformers, we re-ran
the same Gumbel search on Qwen3.5's hybrid architecture, in which $75\%$ of the
token-mixing layers are Gated DeltaNet (linear attention) with non-standard
projections (\texttt{in\_proj\_qkv}, \texttt{in\_proj\_z},
\texttt{linear\_attn.out\_proj}) rather than softmax attention. The same
projection-type hierarchy emerges, and, counter to our worry that an untested
layer family might need \emph{more} precision, the DeltaNet projections turn out
to be the \emph{most} compressible in the model: \texttt{in\_proj\_qkv} selects
\bits{1} for $95.7\%$ of its groups, while the precision-hungry projections are
again the value and MLP up/down projections ($\sim\!70\%$ \bits{1},
$\sim\!30\%$ \bits{2}). The ranking is stable across checkpoints and no group
demands $4$/\bits{8} by mid-training. The practical payoff is that the recipe
needs \emph{no} architecture-specific retuning: extending it to the new
projection names and placing DeltaNet layers at \bits{2} is, if anything,
conservative. This discovery run fixes the \emph{layout}; its \emph{quality} is
then validated by the controlled $1.0$B from-scratch comparison against a matched
FP16 control (Section~\ref{sec:limits}), and the same packed model supplies the
Qwen3.5 inference measurements (Section~\ref{sec:inference}).

\subsection{From adaptive to prescribed, and ``bigger-but-smaller''}\label{sec:bigger}
\begin{figure}[t]
  \centering
  \begin{subfigure}{0.49\textwidth}
    \includegraphics[width=\textwidth]{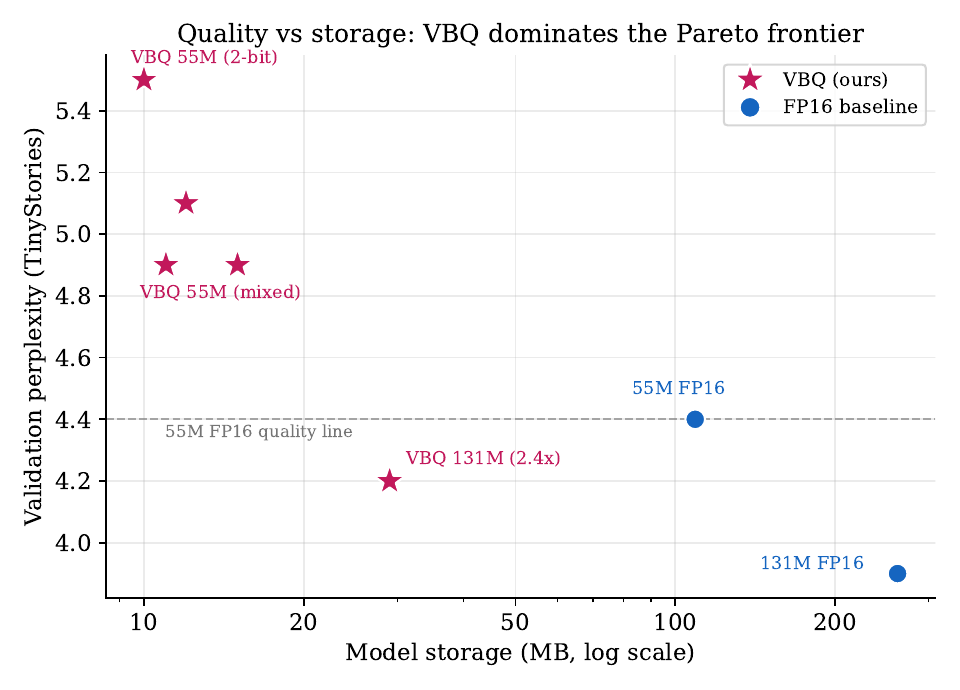}
    \caption{Quality vs.\ storage frontier.}
    \label{fig:pareto}
  \end{subfigure}\hfill
  \begin{subfigure}{0.49\textwidth}
    \includegraphics[width=\textwidth]{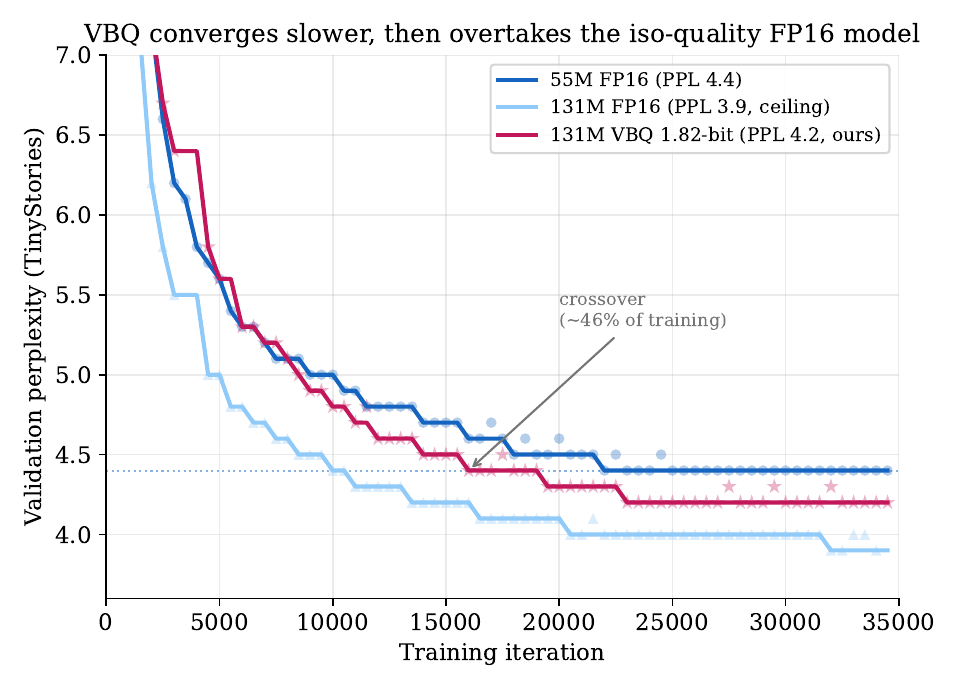}
    \caption{Training crossover (real eval logs).}
    \label{fig:crossover}
  \end{subfigure}
  \caption{\textbf{Bigger-but-smaller.} (a) Validation perplexity vs.\ packed
  storage on TinyStories: every \vbq{} model (crimson stars) sits on or below
  the FP16 baselines (blue circles); the $131$M \vbq{} model at $29$\,MB beats
  the $55$M FP16 model at $109$\,MB. (b) Validation perplexity over training for
  three fixed-architecture $35$K-iteration runs. Faint markers are the raw
  per-eval points; solid lines trace the best perplexity reached so far (a
  running minimum), which is robust to the transient single-eval spikes visible
  in the scatter. The $131$M \vbq{} model ($1.82$ bits) converges more slowly
  than FP16 but keeps improving after the $55$M FP16 model plateaus, crossing its
  quality line (dotted) at $\sim\!46\%$ of training and finishing at PPL $4.2$
  vs.\ $4.4$.}
\end{figure}

Table~\ref{tab:scaling} shows the TinyStories scaling ladder. Three points stand
out. First, \emph{the search is a one-time cost, not a per-model one}: the
discovered layout, frozen into the fixed recipe, equals or beats the adaptive
pilot with no Gumbel machinery at all. This is the practical crux of the
method: the expensive part is \emph{discovering where the bits should go}, and
that needs to happen only once. Once the allocation pattern is known, any new
model can be trained at low precision by simply reading off the recipe (a
per-projection bit map plus a token-frequency table for the head); no
search, no relaxation, no extra hyper-parameters at train time. Every result
below the pilot row in Table~\ref{tab:scaling}, and the entire $1.46$B FineWeb-Edu
scale-up, uses the prescribed recipe directly rather than re-running the search.
Second, the
\emph{frequency-tiered head is the single most impactful component}: switching
the head from uniform \bits{1} (008a) to frequency-tiered (008f) improves PPL by
$0.6$ for only $+0.09$ mean bits. Third, and headline: scaling the model up
$2.4\times$ while staying at $1.82$ bits yields PPL $4.2$, which \emph{beats} the
55M FP16 baseline (PPL $4.4$) at $3.8\times$ less storage
(Figure~\ref{fig:pareto}). Every \vbq{} point lies on or below the FP16 Pareto
frontier.

\begin{table}[t]
  \centering
  \footnotesize
  \caption{\textbf{TinyStories scaling.} A 131M \vbq{} model at $1.82$ bits beats
  a 55M FP16 model on perplexity at $3.8\times$ less storage. ``$\Delta$ PPL'' is
  relative to the 55M FP16 baseline.}
  \label{tab:scaling}
  \begin{tabular}{lrrrrr}
    \toprule
    Model & Params & Mean bits & Val PPL & Storage & $\Delta$ PPL vs.\ 55M FP16 \\
    \midrule
    55M FP16 (baseline)        & 55M  & 16.0 & 4.4 & 109\,MB & n/a \\
    131M FP16 (quality ceiling)& 131M & 16.0 & 3.9 & 262\,MB & $-11\%$ \\
    \midrule
    \vbq{} adaptive pilot      & 55M  & 1.78 & 5.1 & 12\,MB  & $+16\%$ \\
    \vbq{} uniform \bits{2}    & 55M  & 1.50 & 5.5 & 10\,MB  & $+25\%$ \\
    \vbq{} mixed head          & 55M  & 1.59 & 4.9 & 11\,MB  & $+11\%$ \\
    \vbq{} $1.3\times$ scale   & 73M  & 1.66 & 4.9 & 15\,MB  & $+11\%$ \\
    \textbf{\vbq{} $2.4\times$ scale} & \textbf{131M} & \textbf{1.82}
      & \textbf{4.2} & \textbf{29\,MB} & $\bm{-5\%}$ \\
    \bottomrule
  \end{tabular}
\end{table}

\paragraph{Quantization as regularization.}
Figure~\ref{fig:crossover} plots the real validation curves. \vbq{} converges
\emph{slower} than FP16 but keeps improving after FP16 plateaus, crossing the
55M-FP16 quality line near $46\%$ of training. We attribute this to the low-bit
constraint acting as a regularizer: with dropout $0$, the 131M \vbq{} model shows
a smaller train/val gap than its FP16 twin. A practical consequence is that
\vbq{} runs must not be judged early.

\paragraph{Scaling to 1.46B.}
On FineWeb-Edu with the Qwen tokenizer, the same recipe applied to a
$1.46$B-parameter model ($1.77$ mean bits) reaches a best validation loss of
$2.964$ (PPL $19.4$), matching the $593$M FP16 control's best PPL of $19.4$.
The match holds at $\sim\!3.7\times$ less storage
($0.32$ vs $1.19$\,GB) with $2.5\times$ more parameters. The recipe keeps the
frequency-tiered head (mean $1.18$ bits) here too, and its importance is
quantified by the TinyStories ablation above: replacing it with a uniform
\bits{1} head costs $0.6$ PPL for a saving of only $+0.09$ mean bits. What
matters is not the head's \emph{average} bit budget but \emph{where} those bits
go.

\begin{figure}[t]
  \centering
  \includegraphics[width=0.66\textwidth]{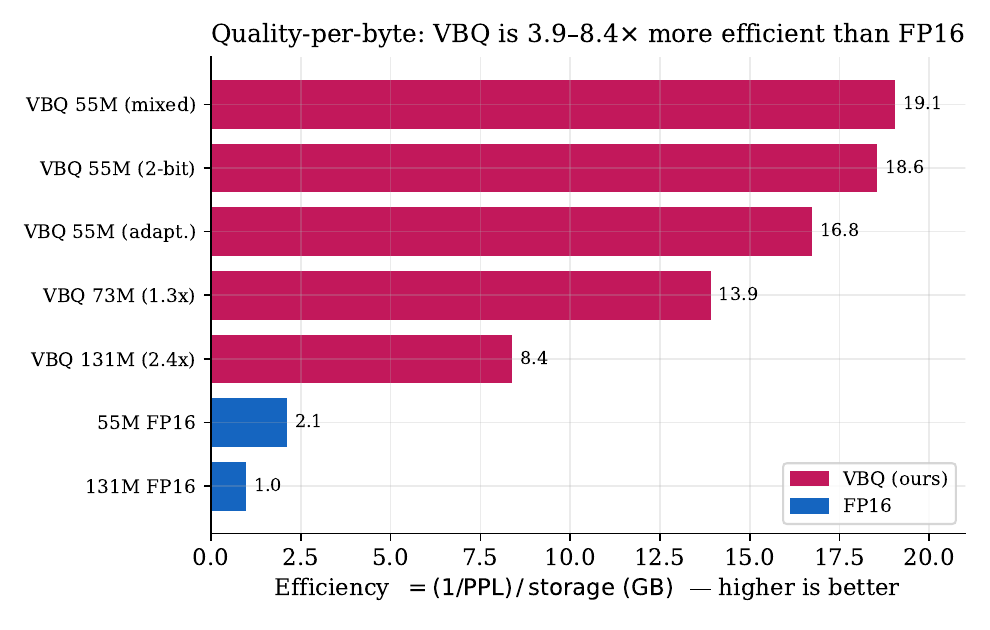}
  \caption{\textbf{Quality-per-byte} $(1/\text{PPL})/\text{storage\,(GB)}$.
  \vbq{} models (crimson) and FP16 baselines (blue) do not overlap; the
  iso-parameter 131M comparison is $8.4\times$.}
  \label{fig:eff}
\end{figure}

\paragraph{Quality-per-byte.}
Summarizing with an efficiency metric $(1/\text{PPL})/\text{storage\,(GB)}$,
\vbq{} models are $3.9$--$8.4\times$ more efficient than FP16 across the ladder
(Figure~\ref{fig:eff}); the mixed-head 55M model is the most efficient point
overall, and there is no overlap between the \vbq{} and FP16 clusters.

\FloatBarrier
\subsection{Distributional analysis: depth self-heals quantization error}\label{sec:kld}
Perplexity can hide behavioral change~\citep{dutta2024accuracy}. We therefore
measure mean token-level KL divergence and argmax-flip rate of the 131M \vbq{}
model against a 131M FP16 reference. The full model has mean KLD $0.21$ nats and
an $18.7\%$ flip rate: at roughly one position in five \vbq{} selects a different
top-1 token, a behavioral cost invisible to PPL, which we report for honesty.

\begin{figure}[t]
  \centering
  \includegraphics[width=0.66\textwidth]{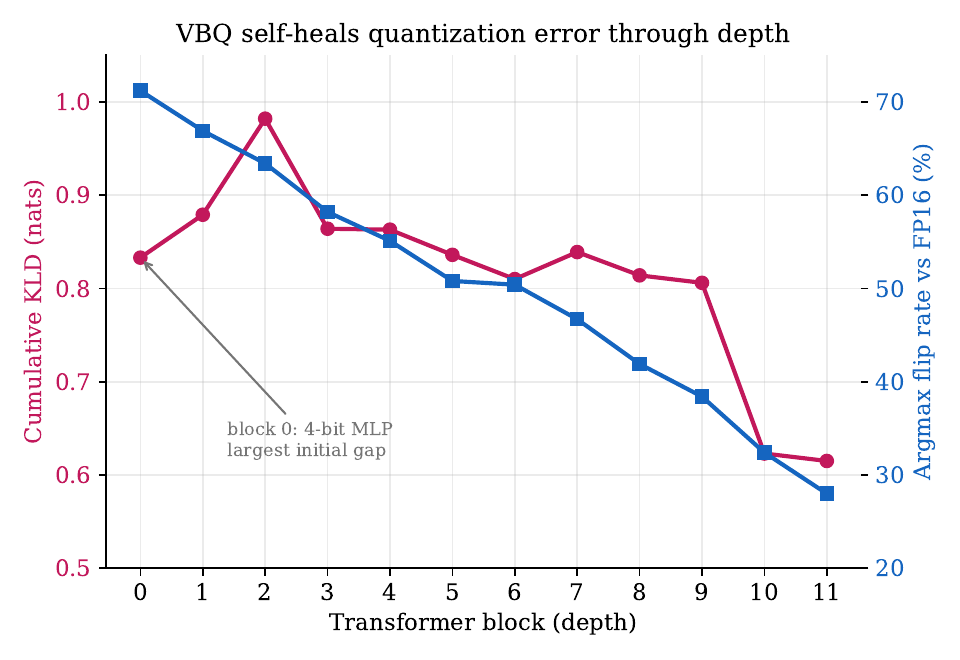}
  \caption{\textbf{Self-healing through depth.} Cumulative KL divergence to FP16
  (crimson) and argmax-flip rate (blue) measured by projecting each block's
  hidden state through a shared head. The early 4-bit block injects the largest
  gap; later 2-bit blocks progressively correct it.}
  \label{fig:selfheal}
\end{figure}

Two findings refine the picture. First, swapping components between the \vbq{}
and FP16 models makes things \emph{worse} (an FP16 head on \vbq{} blocks yields
KLD $0.62$, $3\times$ the full \vbq{} model), showing the blocks and head
\emph{co-adapt}: the low-bit head works precisely because the blocks learned to
produce representations robust to sign-only quantization. Second, a depth probe
that projects each block's hidden state through a single shared FP16 head
(excluding \vbq{}'s own co-adapted head, so blocks are comparable across depth)
shows the representational gap is injected early: the $4$-bit first block
contributes the largest KLD ($0.83$, $71\%$ flips) and is then
\emph{progressively reduced by later $2$-bit blocks}, the probe's flip rate
falling monotonically from $71\%$ to $28\%$ at the final block
(Figure~\ref{fig:selfheal}). This probe sits above the full model's $18.7\%$
end-to-end flip rate precisely because it bypasses \vbq{}'s co-adapted head;
restoring that head recovers the $18.7\%$. \vbq{} networks effectively learn an
error-correcting code over depth. This also explains a negative ablation: adding
precision to individual blocks barely moves KLD ($<\!4.4\%$ for $+0.34$ bits),
because the co-adaptation absorbs whatever precision it is given, making
\emph{more blocks}, not more bits per block, the natural lever.

\subsection{Depth vs.\ width}\label{sec:depth}
The self-healing mechanism suggests depth should be a cheap quality lever, since
each extra $2$-bit block is nearly free in storage. We tested this directly at
fixed width ($d{=}512$): adding depth helps consistently but
\emph{sub-linearly}. A 12-layer model improves PPL by $5$--$6\%$ over 8 layers
($4.9\!\to\!4.6$), but a 26-layer model plateaus at $4.77$, far short of the
$\sim\!4.3$ a naive depth extrapolation predicts, and worse per parameter than
going \emph{wider}. Pure depth is therefore not a substitute for width: the
headline $131$M result wins by being both wider \emph{and} deeper, not deeper
alone.

\FloatBarrier
\section{Inference: storage savings become speed}\label{sec:inference}
A learned bit budget is only useful if it survives to deployment. Because
autoregressive decode at batch size~1 is \emph{memory-bandwidth bound} (each
generated token must stream every weight from memory through the arithmetic
units), a model that stores its weights in fewer bits also \emph{reads} fewer
bytes per token, and decodes faster in direct proportion, provided the weights
are never expanded to FP16 in memory.

\begin{figure}[t]
  \centering
  \includegraphics[width=0.68\textwidth]{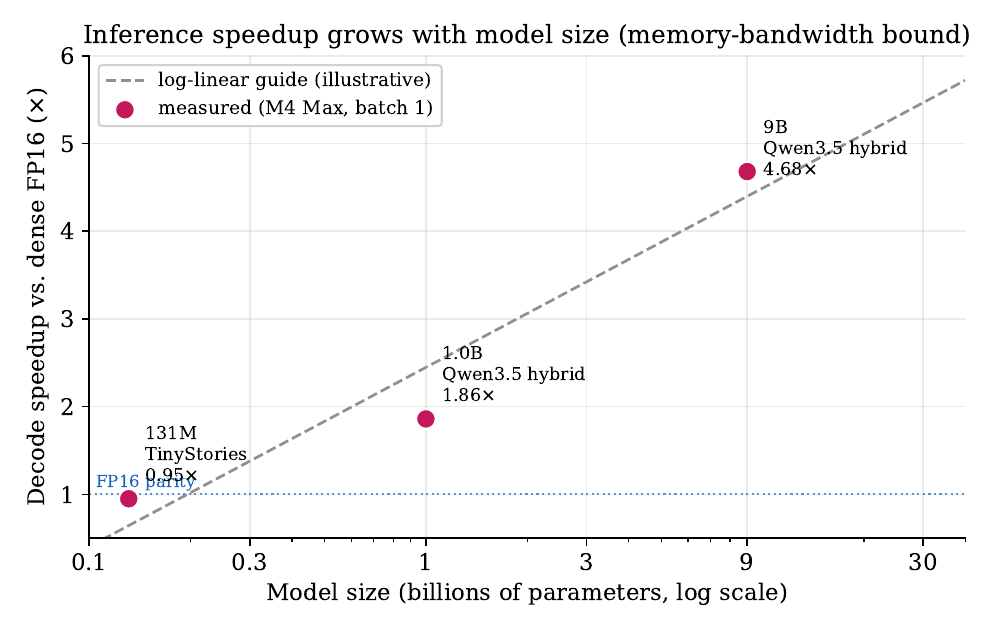}
  \caption{\textbf{The inference win compounds with scale.} Measured
  single-stream decode speedup of the packed \vbq{} model over the same model
  dequantized to dense FP16 (Apple M4 Max, batch~1), plotted against parameter
  count on a log axis. At $131$M the model is too small for weight traffic to
  dominate and \vbq{} runs at FP16 parity ($0.95\times$); by $1.0$B it is
  $1.86\times$ and by $9$B it is $4.68\times$ faster. The dashed line is an
  illustrative log-linear guide through the three measured points (not a fitted
  scaling law), included only to show the direction of the trend. The mechanism
  is memory bandwidth: as models grow, decode spends proportionally more time
  streaming weights, so reading $\sim\!6\times$ fewer bytes per token converts
  more directly into wall-clock speed. This suggests that the combination
  \vbq{} targets (more parameters, $\sim\!3.7\times$ smaller storage, \emph{and}
  faster decode) becomes more attractive precisely at the larger scales where
  inference cost matters most, though we caution that we measure only up to $9$B
  and do not claim a specific extrapolated multiplier beyond that range.}
  \label{fig:inference_scale}
\end{figure}

\paragraph{Fused kernels over packed weights.}
We implement custom fused \emph{dequantize-and-multiply} GEMV kernels (Metal for
Apple silicon, CUDA for NVIDIA) with a specialization per bit bucket
$\{1,2,4,8\}$. Each kernel loads the packed low-bit bytes for a group, expands
them to FP16 \emph{in registers} using the per-group scale, and accumulates the
matrix--vector product, so the full-precision weight matrix is never
materialized in memory and the bandwidth saving is realized end to end. The
mixed-precision head is stored as a small number of frequency-grouped row
buckets.

\paragraph{The speedup grows with model size.}
Because the benefit is bandwidth-driven, it appears only once weight traffic
dominates; i.e., it \emph{scales with model size}
(Figure~\ref{fig:inference_scale}). Table~\ref{tab:inference}
reports the same-weights comparison (identical model and runtime; the FP16
baseline is produced by dequantizing the VBQ weights into dense FP16, so only
the weight representation differs) at batch~1 on an Apple M4 Max, as the median
of repeated runs. At $131$M the model is small enough that per-token overhead,
not bandwidth, dominates, so \vbq{} runs at parity ($0.95\times$) and the win is
purely the $\sim\!2\times$ smaller memory footprint. By $1.0$B decode is
$1.86\times$ faster, and by $9$B it is $4.68\times$ faster ($135$ vs.\ $29$
tok/s), while also fitting in $3.8$\,GB instead of $17$\,GB of peak memory,
i.e.\ running comfortably where the FP16 model is far heavier. The $1.0$B/$9$B
runs use mlx-lm's \texttt{generate\_step} (the real \texttt{async\_eval} decode
path); the $131$M run uses our fused-kernel greedy decode. \vbq{} reproduces the
dense-FP16 model's outputs to numerical parity (verified separately: Pearson
correlation $1.00000$, identical top-1 token), so these are speedups at equal
output, not an accuracy/speed trade.

\begin{table}[t]
  \centering
  \footnotesize
  \caption{\textbf{Compressed decode on an Apple M4 Max} (batch~1, median of
  repeats). The FP16 baseline is the \emph{same model} with its quantized linear
  layers dequantized to dense FP16, isolating the cost of the packed-vs-dense
  weight representation (the token embedding is left quantized in both).
  ``Linear weights'' is the byte size the decode kernels must stream; decode is
  memory-bandwidth bound, so the speedup tracks that saving and grows with model
  size.}
  \label{tab:inference}
  \begin{tabular}{llrrr}
    \toprule
    & & Linear weights & Peak memory & Decode tok/s \\
    Model & Params & FP16\,$\to$\,VBQ & FP16\,$\to$\,VBQ & FP16\,$\to$\,VBQ\ (speedup) \\
    \midrule
    TinyStories LM & 131M & $324\to117$\,MB & $350\to180$\,MB & $592\to564$\quad($0.95\times$) \\
    Qwen3.5 hybrid & 1.0B & $1503\to240$\,MB & $1845\to592$\,MB & $232\to432$\quad($\bm{1.86\times}$) \\
    Qwen3.5 hybrid & 9B   & $15859\to2528$\,MB & $17052\to3783$\,MB & $29\to135$\quad($\bm{4.68\times}$) \\
    \bottomrule
  \end{tabular}
\end{table}

The $9$B row benchmarks decode speed only; the quality of that particular
from-scratch run is a separate, unresolved matter
(Section~\ref{sec:limits}), but the packed model is architecturally
representative, so it is a valid measurement of how the inference benefit scales.

\paragraph{Granularity is deployment-friendly.}
We quantize \emph{contiguous} $64$-weight groups (rather than per-channel or
per-tensor) because a single shared scale over a short contiguous run is what
creates a local quality/precision trade-off (Section~\ref{sec:method}) while
still packing into aligned byte boundaries. Two practical details keep the
packed model fast. First, the deployed artifact is the \emph{recipe}, which is
near-uniform per projection type (mostly \bits{2}, a \bits{4} first block, a
tiered head) rather than a fully heterogeneous per-group map, so a kernel
dispatches one bit bucket per projection, not one per group. Second, the fastest
decode GEMV requires the contracted dimension to be a multiple of $512$; we
zero-pad MLP intermediates to satisfy this (the padded channels are
mathematically inert: $\mathrm{silu}(0)\cdot u = 0$), recovering a $1.67\times$
kernel speedup that would otherwise be lost on awkward shapes such as Qwen's
$8960$-wide MLP. An export-time check validates this contract.

\FloatBarrier
\section{Limitations and Open Problems}\label{sec:limits}
We state the method's scope plainly, and separate what we have \emph{shown}
from what remains \emph{open}.

\paragraph{Economical scaling beyond 1.5B is the main open problem.}
Our positive quality results run from 55M to $1.46$B parameters; economically
validating the recipe beyond that remains open. A separate controlled $1.0$B
gate run on Qwen3.5's hybrid architecture ($1006.7$M total parameters) used a
different tokenizer, a matched FP16 control, and
$\geq\!99\%$ quantization coverage, but was deliberately short at $0.37$
tokens/parameter, about $4\times$ fewer than the $\sim\!1.4$ used in the matched
FineWeb runs. This was by design: it is a cheap \emph{transfer probe} to check
that the recipe carries to a hybrid architecture at all, not a run to
convergence; full matched runs at this scale are precisely the economics we flag
as unsolved. Even so, at the stopped checkpoint \vbq{} was $+0.28$ nats behind
FP16 ($1.33\times$ PPL) with the gap still narrowing over the final third
($0.36\!\to\!0.28$ nats), consistent with \vbq{}'s known slower convergence and
showing no sign of a scaling ceiling. This $1.0$B
gate is not numerically comparable to the $1.46$B FineWeb-Edu standard-transformer
result (architecture, tokenizer, and token budget all differ), so it probes
recipe transfer to a harder hybrid setting rather than contradicting the
standard-transformer parity result. We also attempted a
single $9$B from-scratch knowledge-distillation~\citep{hinton2015distilling}
run, but it was confounded and we do not draw conclusions from it: a coverage bug left $18\%$ of parameters (non-standard
hybrid-attention projections) unquantized so the realized precision was not the
intended $1.77$ bits, and it was trained at only $0.27$ tokens/parameter (about
$5\times$ under the $\sim\!1.4$ used by our successful runs) on an untested
architecture family. A properly-tokened $9$B run (estimated $\sim\!12$B tokens)
was beyond our compute budget. Nothing in our results suggests the method
\emph{stops} scaling; closing the $\ge\!9$B regime economically, and clearing
the strong-baseline bar at that scale (a $4$-bit PTQ of a same-storage teacher),
is future work rather than a counter-result.

\paragraph{Frozen-base depth-boosting does not add useful capacity.}
Motivated by self-healing, we tried to \emph{add} low-bit correction blocks onto
a converged 12-layer model with the base frozen (a gradient-boosting analogy with
near-identity initialization). Training loss improved but validation did not
(a classic overfit signature, with intermittent instability); the added blocks
fit batch idiosyncrasies rather than generalizable residual signal. At least with
a small frozen base, post-hoc correction depth is not a shortcut to the
from-scratch win. We suspect a larger pretrained base would \emph{not} rescue
this either: the ``bigger-but-smaller'' win comes from weights co-adapting to low
bit-widths \emph{during} training, which a frozen or otherwise pretrained base
cannot inherit; and indeed, separate attempts to search per-group bit-widths
post-hoc on \emph{frozen} pretrained weights were dominated by trivial uniform
4-bit. The more promising direction is therefore to scale the from-scratch
``bigger-but-smaller'' recipe itself, not to adapt pretrained bases.

\paragraph{Scope.}
Our positive results are on English next-token modeling up to $1.5$B parameters,
on two corpora, with perplexity and KL/flip metrics; we do not report downstream
task suites (MMLU/GSM8K/etc.) at scale. The method targets the
\emph{train-from-scratch} setting; it is not, on current evidence, a superior
\emph{post-training} compressor competing with GPTQ, AWQ, QuIP\#, or AQLM at
$4$ bits.

\section{Conclusion}
\vbq{} treats numerical precision as a per-group, learnable resource rather than
a global constant. A Gumbel-Softmax relaxation, made trainable by gradient
isolation and alternating optimization, discovers a strongly heterogeneous
bit-width hierarchy that is stable enough to freeze into a reusable recipe. The
central lesson is that spending a fixed bit budget \emph{unevenly} is materially
better than spending it uniformly: with the recipe, models become simultaneously
\emph{larger} in parameter count and \emph{smaller} in storage than their FP16
equivalents, and $3.9$--$8.4\times$ more efficient in
quality-per-byte. This is consistent with the $\sim\!4$-bit compression floor
that recurs across independent settings~\citep{dettmers2023case,apple2024fm,
searchio2022neuralhashing}: \vbq{} reaches that floor not by rounding each weight
to four bits, but by redistributing a comparable budget across more,
lower-precision weights. This is not only a storage result: the same packed weights
decode faster through fused low-bit kernels, increasingly so with scale (up to
$4.7\times$ at $9$B), because decode is memory-bandwidth bound. The three wins
compound rather than trade off: a \vbq{} model can hold \emph{more} parameters,
occupy \emph{about a quarter} of the storage, and decode \emph{several times
faster} than its FP16 counterpart at once, and because the speedup is
bandwidth-driven it grows with model size, so the combination is most valuable
exactly where serving cost dominates. A KL-divergence analysis reveals \emph{why} the
aggressive bit budget is survivable: depth self-heals the error injected by early
layers, so the network effectively learns an error-correcting code that lets a
non-uniform allocation pay off. The remaining frontier is to push the
from-scratch ``bigger-but-smaller'' recipe economically past $1.5$B parameters.
Precision should not be treated merely as a post-training tuning knob, but as a
structural design input to training itself: choosing the bit layout up front
yields faster, more efficient models that can operate closer to the quality/size
Pareto frontier out of the box.

\paragraph{Reproducibility.}
All figures are regenerated from the released training logs and tabulated
results by the script in \texttt{paper/figures/}.

\small
\bibliographystyle{plainnat}
\bibliography{references}

\end{document}